\pdfoutput=1
\documentclass[10pt, a4paper]{article}

\usepackage{lrec-coling2024} % this is the new style
\usepackage{CJKutf8}
\usepackage{amsmath}
\usepackage{multirow}
\usepackage{hyperref}

\usepackage{natbib}
\usepackage{multibib}
\usepackage{multirow, booktabs}

\makeatletter
\def\@mb@citenamelist{cite,citep,citet,citealp,citealt,citepalias,citetalias}
\makeatother
\newcites{languageresource}{~}

\usepackage{graphicx}
\usepackage{tabularx}
\usepackage{soul}

\usepackage{xcolor}
\usepackage{hyperref}
 \definecolor{darkblue}{rgb}{0, 0, 0.5}
  \hypersetup{colorlinks=true, citecolor=darkblue, linkcolor=darkblue, urlcolor=darkblue}

\usepackage{xstring}

\usepackage{color}

\title{CMNER: A Chinese Multimodal NER Dataset based on Social Media}

\name{Yuanze Ji, Bobo Li, Jun Zhou, Fei Li, Chong Teng, Donghong Ji} 

\address{Key Laboratory of Aerospace Information Security and Trusted Computing, Ministry \\ of Education, School of Cyber Science and Engineering, Wuhan University, China \\
         \{yz\_in\_whu, lifei\_csnlp\}@whu.edu.cn} 
\abstract{
Multimodal Named Entity Recognition (MNER) is a pivotal task designed to extract named entities from text with the support of pertinent images. Nonetheless, a notable paucity of data for Chinese MNER has considerably impeded the progress of this natural language processing task within the Chinese domain. Consequently, in this study, we compile a Chinese Multimodal NER dataset (CMNER) utilizing data sourced from Weibo, China's largest social media platform. Our dataset encompasses 5,000 Weibo posts paired with 18,326 corresponding images. The entities are classified into four distinct categories: person, location, organization, and miscellaneous. We perform baseline experiments on CMNER, and the outcomes underscore the effectiveness of incorporating images for NER. Furthermore, we conduct cross-lingual experiments on the publicly available English MNER dataset (Twitter2015), and the results substantiate our hypothesis that Chinese and English multimodal NER data can mutually enhance the performance of the NER model. We will release CMNER at \url{https://github.com/Jyz99/CMNER}
 \\ \newline  }

\begin{document}

\maketitleabstract

\section{Introduction}
Named entity recognition (NER), as a fundamental task in natural language processing (NLP), has been greatly explored and applied to a wide range of other NLP subtasks, including relation extraction \cite{wei2019novel,li2021mrn}, entity linking \cite{le2018improving}, and so on. While previous NER approaches \cite{li2021span,li2022unified} focused on textual data have achieved significant success, real-world scenarios often present data in a multimodal manner. Consider social media platforms as an example, user-generated content typically integrates text and images, rich with various kinds of named entities. Such scenarios challenge the efficacy of conventional text-based NER methods, underscoring the imperative for multimodal named entity recognition (MNER).

\begin{figure}[!t]
    \centering
    \includegraphics[width=0.75\linewidth]{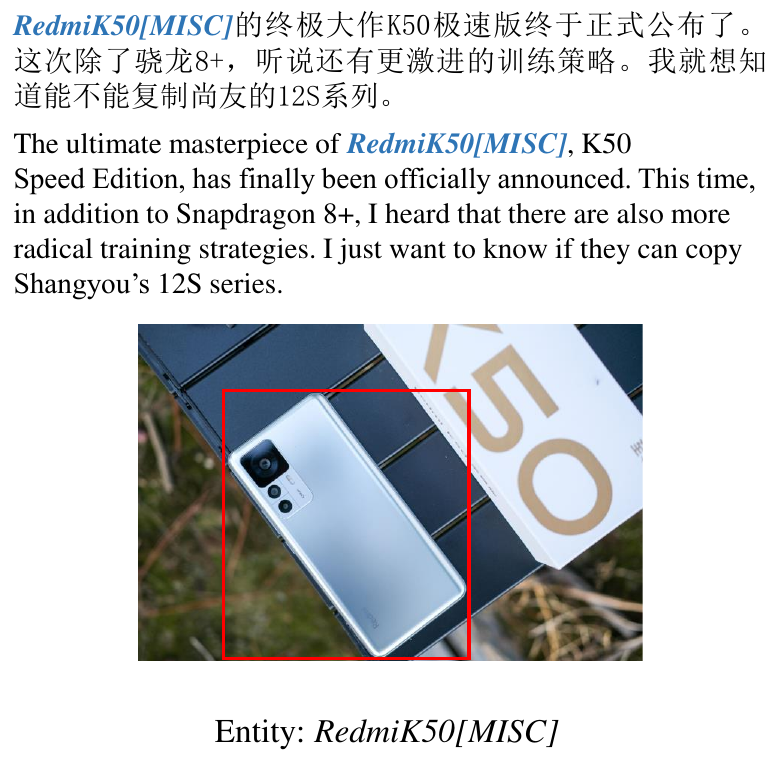}
    \caption{An example of multimodal Weibo post.}
    \label{1}
\end{figure}

Compared to NER, MNER takes not only the text but also relevant images as input, allowing it to digest copious underlying visual information to assist in the detection and classification of entities within the text. Figure \ref{1} provides an illustration of a multimodal Weibo post example. Under the paired image, we can easily discern that ``RedmiK50'' refers to a mobile phone, suggesting it should be categorized as a MISC entity. However, we might only identify ``Redmi'' as the name of an organization solely based on the text. In order to evaluate the impact of images on NER, two widely-utilized English MNER datasets, Twitter2015 \cite{zhang2018adaptive} and Twitter2017 \cite{yu2020improving}, were introduced, giving rise to numerous research endeavors \cite{moon2018multimodal,zhang2021multi,wang2021ita}. Nonetheless, there remains a dearth of comprehensive Chinese MNER datasets, which could potentially hinder the development of MNER in the Chinese context.

Based on our newly constructed CMNER dataset, we execute the CMNER task following the methodologies outlined in \cite{zhang2018adaptive} as well as \cite{yu2020improving}, presenting the results as baselines. Concretely, we conduct two benchmark experiments. The first designs an adaptive co-attention network to generate word-aware visual representations established on CNN-BiLSTM-CRF, while the second employs a unified multimodal transformer to facilitate the final entity predictions. The results not only demonstrate that the integration of images indeed promotes NER performance, leaving space for further refinement, but also validate the efficacy and high quality of our CMNER dataset. In addition to the aforementioned work, we explore the potential mutual contributions of English and Chinese towards advancing NER performance. We conduct a series of cross-lingual experiments utilizing the well-known Twitter2015 dataset and our newly proposed CMNER dataset to corroborate this conjecture.

To give a concise overview, the main contributions of this paper can be summarized as follows:

\begin{itemize}
\item We introduce a completely new, manually annotated, high-quality Chinese multimodal NER dataset derived from Chinese social media. To the best of our knowledge, it is the first dataset that accurately emulates the one-text-multi-image characteristic of Weibo posts.

\item Benchmark results are provided based on our CMNER dataset, demonstrating the potential for further enhancements on CMNER and offering foundational performance metrics for future research endeavors.

\item We conduct a sequence of cross-lingual experiments using Twitter2015 and CMNER, confirming the reciprocal function to enhance NER performance between Chinese and English.
\end{itemize}

\section{Related Work}
\subsection{NER}

NER has long stood as a foundational task in natural language processing, owing to its extensive applicability. Current NER approaches can be categorized into four main types: Sequence labeling methods conceptualize NER as a task of assigning tags to each word in a sequence \cite{ma2016end}. Span-based approaches treat NER as a span classification task \cite{sohrab2018deep}, overcoming nested NER. The hypergraph model serves as other methods to tackle overlapped NER \cite{lu2015joint,katiyar2018nested,wang2018neural}. Generation approaches view NER as a sequence generation task \cite{lu2022unified,zhang2022bias}, establishing a unified framework capable of handling both flat and nested NER.

Recently, the task has been extended to more practical scenarios, for instance, the few-shot NER \cite{ding2021few} and the cross-domain NER \cite{liu2021crossner}. To address these challenges, approaches like data augmentation \cite{chen2021data, zhou2021melm} and prototype networks \cite{huang2021few, ding2021few} have been employed. In addition, \cite{shen2023promptner} designs a dual-slot multi-prompt template to facilitate entity detecting and typing. 

\subsection{Multimodal NER}

With the thriving development of social media platforms, multimodal named entity recognition (MNER) has garnered the attention of researchers, leveraging visual inputs to enhance NER. The alignment and fusion of textual and visual features constitute the crux of MNER. Some studies \cite{moon2018multimodal,zheng2020object,lu2018visual} have proposed diverse cross-modal attention mechanisms, while others have introduced the latest neural network architectures to MNER, such as Transformer-based approaches \cite{chen2022hybrid,xu2022maf}, modality translation-based approaches \cite{chen2020can}, graph neural network-based approaches \cite{zhang2021multi,zhao2022learning}, and prompt-based approaches \cite{wang2022promptmner}. Besides, \cite{jia2023mner} has devised an end-to-end Machine Reading Comprehension (MRC) framework for MNER with query grounding, which utilizes queries to enhance the MNER process.

\subsection{Cross-lingual Transfer}

Present cross-lingual transfer learning can be broadly classified into two main types: model transferring and annotation projection. The former attempts to train a model on a source language corpus and subsequently adapt it to the target language \cite{tiedemann2015improving,wu2020single}. The latter employs existing annotated corpus in the source language to generate a target language version through translation and projection, and then train a model on the target language corpus \cite{daza2019translate,fei2020cross}.

For cross-lingual NER, prevalent methods fall into three categories: Feature-based methods capture language-independent features for cross-lingual transfer \cite{keung2019adversarial,wu2019beto}. Translation-based methods involve translating annotated source-language datasets to generate pseudo training data \cite{jain2019entity,liu2021mulda}. Self-training approaches utilize a trained model on annotated source-language data to produce pseudo-labeled data in the target language \cite{chen2021advpicker,li2022unsupervised}. \cite{zhou2023improving} integrates representation learning and pseudo-label refinement to enhance self-training for cross-lingual NER. In this paper, we perform translation on the source-language corpus and implement label mapping to facilitate cross-lingual NER.

\section{Dataset Construction}
\subsection{Dataset Design}

The very first step in this work is to construct an appropriate dataset. Considering the demand for multimodal data, we turn to the biggest Chinese social media platform, Weibo, where posts are composed of texts, pictures, or videos. There is no character limit for Weibo posts, and each post can have 0 to 18 images. To faithfully replicate real-life social media scenarios while adhering to our multimodal requirements, we exclusively gather Weibo posts that simultaneously feature both images and text. In instances where multiple images are present, we collect all of them. 

Following the annotation scheme of CoNll2003, our objective is to identify entities of types PER, LOC, ORG, and MISC in texts. Therefore, we collect Weibo data from the categories of sports, traveling, and technology through the topic in Weibo like ``\#\begin{CJK*}{UTF8}{gbsn}体育\end{CJK*}\#(\#Sports\#)''. Within these categories, we observe distinct entity distributions: the sports category predominantly features PER entities, the traveling category exhibits a higher prevalence of LOC entities, and the technology posts contain more ORG entities. We use a Python crawler to gather data from the Weibo website, during which we set the maximum character limit to 250. Upon obtaining the source data, we first filter out cases involving pornography, violence, or discrimination to ensure its appropriateness. when concerning users' private information, we also redact this portion of the texts to safeguard privacy.

\subsection{Preparation and Annotation}

Before the annotation process, we implement regular expressions and rules to further clean the Weibo texts, which involves removing garbled characters, unnecessary blank space, and so on. Additionally, we engage some annotators to conduct a filtering process on the data. Firstly, we eliminate images that are not directly relevant to the text. Secondly, we filter out data where the text does not include any named entities, guaranteeing that each Weibo post contains at least one named entity.

Following the annotation scheme of CoNll2003, the BIO method is applied to assign labels to each character. We implement a comprehensive annotation protocol to mitigate subjective discrepancies and annotate four common entity types (i.e., PER, LOC, ORG, MISC). Specifically, we only annotate the names of celebrities, while in terms of location, we annotate them down to the finest granularity in the text. The category of organizations encompasses enterprises, sports teams, television stations, and the like. Due to the intrinsic disparities between Chinese and English, the annotation of MISC entities in our dataset primarily concentrates on common proper nouns as well as those within the aforementioned fields, which differs somewhat from that in CoNll2003.

We exploit the open-source text annotation tool, Doccano, to accomplish the labeling process. Before the formal annotation procedure, we randomly select 100 samples for independent annotation by all annotators. When the inter-annotator agreement reaches 90\%, we start the whole annotation process. Each sample is assigned to three annotators. In cases where their annotations diverge, we hold a separate discussion to determine the annotation for that particular piece of data.

\subsection{Data Statistics}

As illustrated in Table \ref{statistic}, the CMNER dataset comprises a total of 5,000 Weibo posts and 18,326 images among which there are 1,725 posts related to sports, 1,737 related to traveling, and 1,538 related to technology. The dataset is randomly partitioned into training, validation, and test sets in a ratio of 3:1:1 while maintaining a roughly similar distribution across each category. Furthermore, the detailed annotation of entity labels is provided. Specifically, the dataset encompasses 9,850 instances of ORG entities, whereas the count of MISC entities only reaches 2,870.

\begin{table}[!t]
\centering
% \small
\resizebox{0.48\textwidth}{!}{
\begin{tabular}{ccccc}
\hline

\hline
             & Train & Dev  & Test & Total \\
             \hline
PER          & 3,814  & 1,313 & 1,287 & 6,414  \\
LOC          & 4,760  & 1,546 & 1,604 & 7,910  \\
ORG          & 6,045  & 1,942 & 1,863 & 9,850  \\
MISC         & 1,754  & 594  & 522  & 2,870  \\
\hline
Total        & 16,373 & 5,395 & 5,276 & 27,044 \\
Num of Weibo & 3,000  & 1,000 & 1,000 & 5,000 \\
\hline

\hline
\end{tabular}
}
\caption{Statistic of the CMNER.}
\label{statistic}
\end{table}

According to the statistics in Table \ref{statistic}, there are 27,044 entities in total in the CMNER dataset. On average, each post is paired with approximately 3.67 images, while the average number of entities per post amounts to about 5.4.

\begin{figure*}[!t]
    \centering
    \includegraphics[width=0.99\linewidth]{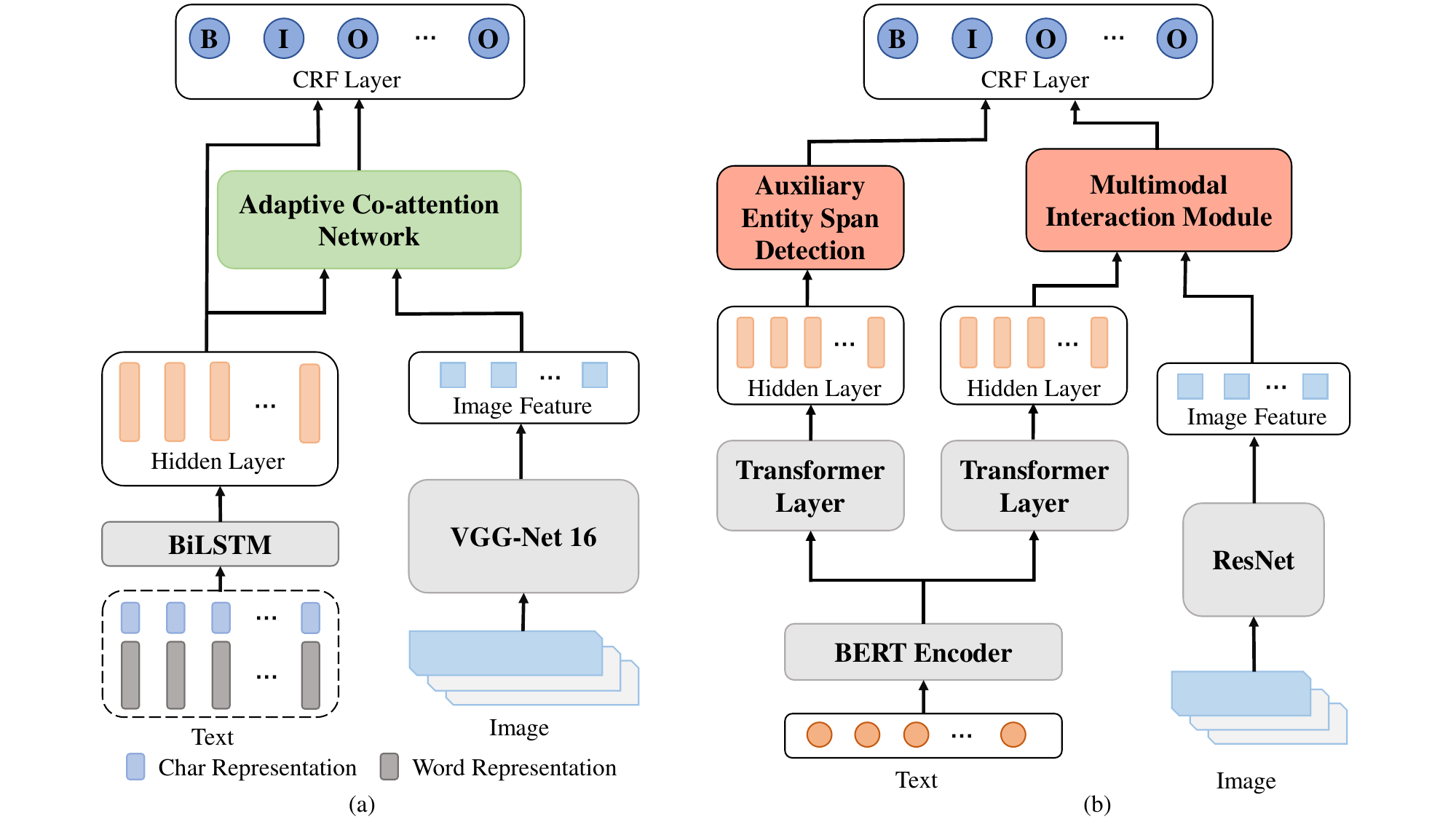}
    \caption{Two baseline models in the experiment. (a) is the architecture of ACN and (b) is the architecture of UMT.}
    \label{model}
\end{figure*}

\section{Baseline Experiments}
In this section, we first delineate the CMNER task and introduce the selected baselines for this study. We then present the results and conduct a comprehensive analysis of the CMNER.

\subsection{Task Formulation}

Compared to NER, multimodal named entity recognition incorporates image information as the auxiliary input to enhance entity recognition. Chinese MNER, a specific instance of MNER, deals with Chinese language texts. Thus we can formulate the CMNER task as follows: Given a Chinese sentence $S$ and its associated image set $V$ as input, the objective is to detect and classify all the entities in the text. The image set $V$ here can include more than one picture.

We view the task as a sequence labeling problem. Given $S =\{s_1\ldots s_n\}$ denotes a sequence of Chinese characters and $V =\{v_1\ldots v_m\}$ represents its relevant images, the output is a label sequence $y =\{y_1\ldots y_n\}$ corresponding to the $S$. $y$$\in$$Y$ and $Y$ = $\{B-x, I-x, O\}$, where $x$$\in$$\{PER, LOC, ORG, MISC\}$.

\subsection{Model Architectures}

To provide reliable baselines for CMNER, we conduct experiments with two fundamental models specifically designed for multimodal named entity recognition and we also perform some ablation studies to demonstrate the impact of incorporating images in the CMNER task.

As depicted in Figure \ref{model} (a), AdaCAN-CNN-BiLSTM-CRF (ACN) \cite{zhang2018adaptive} is an MNER model based on CNN-BiLSTM-CRF, designing an adaptive co-attention network to integrate information from both text and image. It employs 16-layer VGGNet \cite{simonyan2014very} as the image feature extractor and uses CNN and BiLSTM to derive representations for words in the input sentence. Figure \ref{model} (b) illustrates the framework of UMT-BERT-CRF (UMT) \cite{yu2020improving}, another MNER model that leverages Transformer \cite{vaswani2017attention}. It applies a cross-modal attention mechanism on word representations from BERT \cite{devlin2018bert} and visual representations from ResNet \cite{he2016deep}. Additionally, UMT proposes an auxiliary entity span detection module to enhance the MNER process.

Unlike the aforementioned models, multiple images may be present in our CMNER cases. To address this, we first generate embeddings for each image and then compute their average to form the final visual representation for subsequent modules. We also experiment with the prevalent pre-trained language models such as UIE \cite{lu2022unified} and Chinese CLIP \cite{yang2022chinese} to address the problem. However, their performance does not prove to be competitive, which could be attributed to the fact that these models are not pre-trained with a specific focus on the MNER task.

\subsection{Experiment Setup}\label{sec:setup}

% Please add the following required packages to your document preamble:
% \usepackage{multirow}
% \begin{table}[]
% \begin{tabular}{ccccccccc}
% \multicolumn{2}{c}{\multirow{3}{*}{methods}} & \multicolumn{7}{c}{CMNERD}                                         \\
% \multicolumn{2}{c}{}                         & \multicolumn{4}{c}{Single Type (F1)} & \multicolumn{3}{c}{Overall} \\
% \multicolumn{2}{c}{}                         & PER     & LOC     & ORG     & MISC   & P       & R       & F1      \\
% \multirow{2}{*}{ACN}     & text           & 78.62   & 67.15   & 73.80   & 73.87  & 73.37   & 72.55   & 72.93   \\
%                             & text+image     & 78.41   & 68.45   & 75.69   & 75.66  & 76.26   & 72.34   & 74.22   \\
% \multirow{2}{*}{UMT}        & text           & 93.07   & 87.73   & 89.07   & 82.57  & 87.24   & 90.78   & 88.98   \\
%                             & text+image     & 93.67   & 88.12   & 89.70   & 83.08  & 87.92   & 91.14   & 89.50  
% \end{tabular}
% \caption{The results of baselines under only the text inputs and both the text and visual inputs.
% \lifei{双栏表}
% }
% \label{mner_res}
% \end{table}

% Please add the following required packages to your document preamble:
% \usepackage{multirow}
\begin{table*}[!t]
\centering
\resizebox{0.99\textwidth}{!}{
\begin{tabular}{ccccccccc}
\hline

\hline

\multicolumn{2}{c}{\multirow{3}{*}{methods}} & \multicolumn{7}{c}{CMNERD}                                                                                                                                            \\
\multicolumn{2}{c}{}                         & \multicolumn{4}{c}{Single Type (F1)}                                                          & \multicolumn{3}{c}{Overall} \\
\cmidrule(r){3-6}\cmidrule(r){7-9}

\multicolumn{2}{c}{}                         & PER                   & LOC                   & ORG                   & MISC                  & P                     & R                     & F1                    \\
\hline
\multirow{3}{*}{AdaCAN}     & text           & 78.62                 & 67.15                 & 73.80                 & 73.87                 & 73.37                 & \textbf{72.55}                 & 72.93                 \\
                            & text+one image & \textbf{79.02}        & 67.39                 & 75.01                 & 73.55                 & 76.04(+2.67)                 & 71.33(-1.22)                 & 73.62(+0.69)                 \\
                            & text+all images& 78.41                 & \textbf{68.45}        & \textbf{75.69}        & \textbf{75.66}        & \textbf{76.26}(+2.89)        & 72.34(-0.21)        & \textbf{74.22}(+1.29)         \\
                            \hline 
\multirow{3}{*}{UMT}        & text           & 93.07                 & 87.73                 & 89.07                 & 82.57                 & 87.24                 & 90.78                 & 88.98                 \\
                            & text+one image           & \textbf{93.84}                 & \textbf{88.33}                 & 89.34                 & 82.00                 & 87.49(+0.25)                 & \textbf{91.31}(+0.53)                 & 89.36(+0.38)                 \\
                            & text+all images     & 93.67  & 88.12 & \textbf{89.70} & \textbf{83.08} & \textbf{87.92}(+0.68) & 91.14(+0.36) & \textbf{89.50}(+0.52) \\

                            \hline

                            \hline
\end{tabular}
}
\caption{Experimental results of baselines conducted using only textual input, using textual input with one randomly selected image, and using textual input with all available images.}
\label{mner_res}
\end{table*}

In the ACN model, we set the maximum sentence length to 250 characters, and the dimension of character embeddings to 300. Truncation or zero-padding is applied as needed. Characters out of the embedding vocabulary are initialized by randomly sampling from a uniform distribution of [-0.25, 0.25]. The other parameters are configured in accordance with the specifications provided in \cite{zhang2018adaptive}.

Similarly, in the UMT model, we set the maximum input sentence length to 250 and the batch size to 16. We initialize the word embeddings using BERT-base-Chinese and utilize a pre-trained 152-layer ResNet to extract image representations and fix them while training. The number of cross-modal attention heads is set to 12, while the remaining parameters are kept consistent with those outlined in \cite{yu2020improving}.

\begin{figure*}[!ht]
    \centering
    \includegraphics[width=0.99\linewidth]{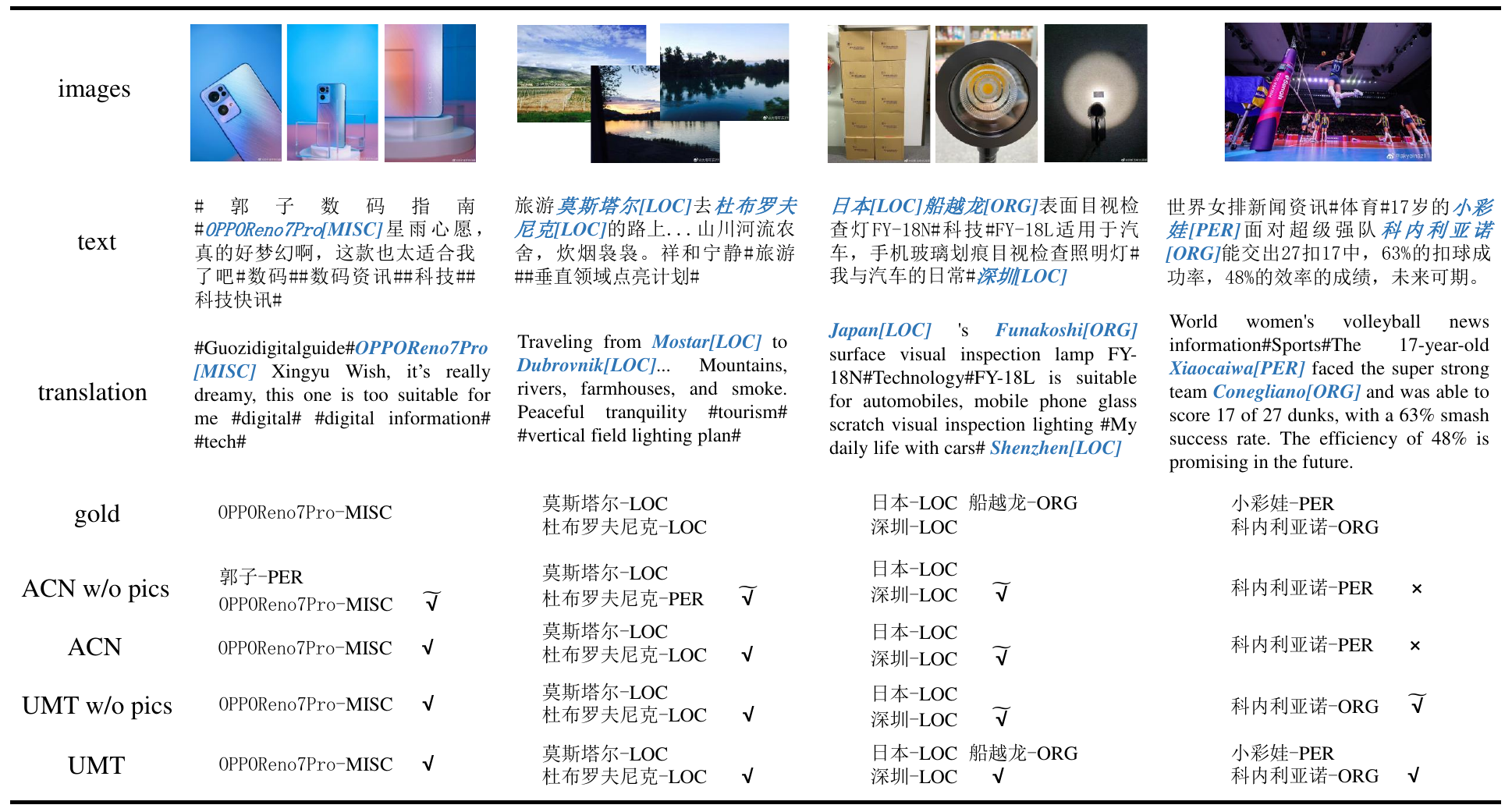}
    \caption{Examples of MNER. The ``gold'' is the golden entity type in the text and ``w/o pics'' means without visual inputs. ``\begin{CJK*}{UTF8}{gbsn}√\end{CJK*}'' denotes the prediction is completely right, ``×'' denotes the prediction is totally wrong, and ``$\tilde{\begin{CJK*}{UTF8}{gbsn}√\end{CJK*}}$'' denotes the prediction is incomplete right.}
    \label{case}
\end{figure*}

We employ precision(P), recall(R), and F1 score(F1) as the evaluation metrics to assess the performance of all entity types. For each individual entity type, we exclusively use the F1 score as the evaluation metric.

\subsection{Results}

Table \ref{mner_res} displays the results of baselines. The upper set of results corresponds to ACN, while the lower set corresponds to UMT. We conduct experiments using only textual input, using textual input with one randomly selected relevant image, and using textual input with all corresponding images for both models. As anticipated, UMT outperforms ACN, achieving an F1 score of 89.50, which is notably higher than ACN's score of 74.22. Even with text-only input, UMT demonstrates superior performance compared to ACN with multimodal inputs. We attribute this to the robust contextual understanding and semantic expression capabilities of the BERT model. Experimental results show that both models achieve the highest F1 score in PER entity recognition, while their performance in the LOC and MISC categories is relatively weaker. This result aligns with the feature that names in Chinese are relatively easier to identify compared to potentially implicit and symbolic representations of locations (e.g., ``\begin{CJK*}{UTF8}{gbsn}仙游\end{CJK*}(Xianyou)'' and so on). We speculate that the lack of distinct characteristics in MISC entities may account for the lower F1 scores.

\begin{figure*}[!t]
    \centering
    \includegraphics[width=0.99\linewidth]{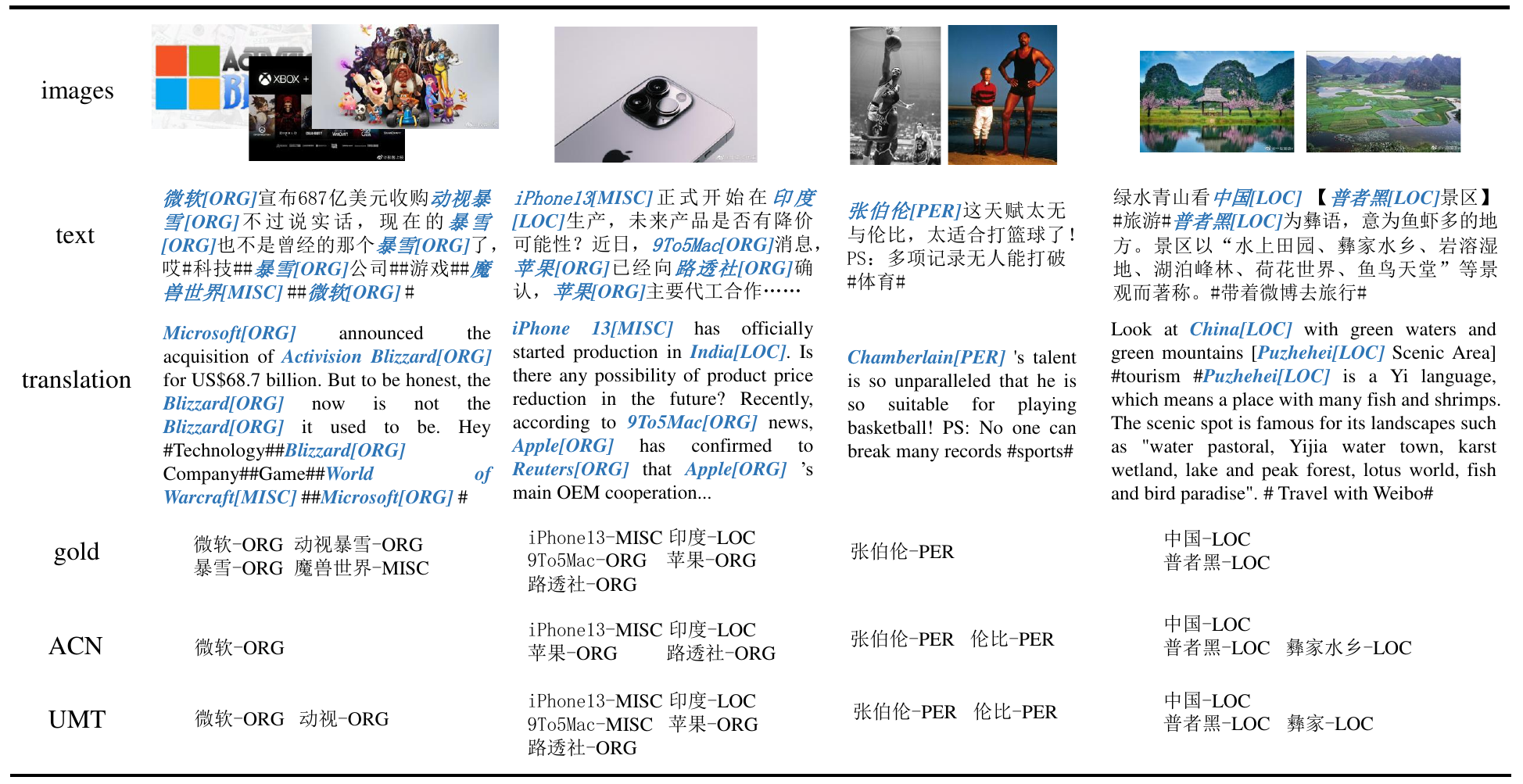}
    \caption{Cases of different error types.}
    \label{error}
\end{figure*}

Table \ref{mner_res} also highlights that the fusion of image information significantly assists in entity recognition and enhances the performance of both two benchmark models. Specifically, with the incorporation of one randomly selected paired image, ACN's F1 score increases from 72.93 to 73.62 across all entity types. Particularly noteworthy are the improvements in the ORG categories, where there is a gain of 1.21 points. The precision of all entity types notably increases from 73.37 to 76.04, whereas, the recall score drops from 72.55 to 71.33, which may be attributed to the noise introduced during the stochastic image selection process. When a relevant image is randomly inputted, UMT exhibits a similar trend, with an enhancement of 0.38 in the overall F1 score, rising from 88.98 to 89.36, and the F1 scores for specific entity categories show improvements as well. UMT's precision and recall scores of all entity types also show an advancement. However, the performance of ACN shows a more significant improvement compared to UMT, and we conjecture that the varying degrees of performance enhancement between the two models may be attributed to the strong contextual comprehension capability of BERT, which may render image features less impactful on its performance.

Furthermore, as depicted in Table \ref{mner_res}, the experimental results for both models demonstrate notable improvements when utilizing textual input in conjunction with all relevant images. Compared to the scenario where only one image is considered, both models exhibit enhanced performance. Specifically, ACN's overall F1 score increases from 73.62 to 74.22, with all entity types (except PER) exhibiting similar improvements. The overall precision and recall scores experience gains of 0.22 and 1.01 points, respectively, which illustrates the effectiveness of incorporating all relevant images for ACN. Likewise, UMT shows a further enhancement in performance and achieves its highest F1 score of 89.50, affirming the efficacy of integrating all pertinent images for UMT. These results not only emphasize the utility of visual inputs in named entity recognition but also validate the necessity and research significance of the one-text-multi-image characteristic in our proposed CMNER dataset.

\subsection{Analysis}

\paragraph{Case Study}

Figure \ref{case} provides four specific examples for further analysis. The left two columns demonstrate the effectiveness of images on ACN while the right two columns illustrate their impact on UMT. In the first case, without visual input, ``\begin{CJK*}{UTF8}{gbsn}郭子\end{CJK*}(Guozi)'' is mistakenly identified as a PER entity. However, when images are introduced, the model can learn that no person appears in the image, leading to a completely accurate prediction. The second example showcases how images can rectify the model's misclassification of entity types. With the assistance of images, ACN correctly identifies ``\begin{CJK*}{UTF8}{gbsn}杜布罗夫尼克\end{CJK*}(Dubrovnik)'' as a LOC entity rather than a PER. The third instance demonstrates that visual inputs also benefit UMT in achieving accurate recognition. When presented with only the text, UMT cannot infer the information that ``\begin{CJK*}{UTF8}{gbsn}船越龙\end{CJK*}(Funakoshi)'' is a certain company's name. The last case involves a piece of women's volleyball news and a picture of an athlete without which the model misses the PER entity ``\begin{CJK*}{UTF8}{gbsn}小彩娃\end{CJK*}(Xiaocaiwa)''. These examples collectively substantiate that introducing auxiliary visual inputs not only enables the model to accurately detect entity spans but also correctly classify them, thereby improving overall performance. We can also observe that UMT makes perfect predictions in the first two examples, which is consistent with the experimental results discussed earlier. 

\paragraph{Error Study}

We conduct further error analysis on both models and Figure \ref{error} lists a few examples. We categorize the errors into four types: (1) inaccurate entity span detection, for example, in the first case, ``\begin{CJK*}{UTF8}{gbsn}动视暴雪\end{CJK*}(Activision Blizzard)'' is an entire ORG entity but UMT only identifies ``\begin{CJK*}{UTF8}{gbsn}动视\end{CJK*}(Activision)'' as an ORG entity; (2) entity misclassification, as in the second example, ``9To5Mac'' is improperly categorized as a MISC entity by UMT; (3) entity omission, for instance, ACN fails to recognize that ``9To5Mac'' is an entity at all; (4) labeling non-entities as entities, as shown in the last two cases, both models mistakenly consider ``\begin{CJK*}{UTF8}{gbsn}伦比\end{CJK*}(Lunbi)'' as a person and ``\begin{CJK*}{UTF8}{gbsn}彝家水乡\end{CJK*}(Yijia water town)'' or ``\begin{CJK*}{UTF8}{gbsn}彝家\end{CJK*}(Yijia)'' as a location. We attribute these errors to three main reasons: (1) fine-grained annotation rules: the annotation rules we formulate are quite detailed, which imposes higher demands on the model; (2) low correlation between images and text: there are cases where the correlation between images and text is low, introducing noise to the model; (3) additional or insufficient image information: pictures may contain extra entities that are not mentioned in the text, and there are also instances where entities mentioned in the text do not appear in the image, which accounts for the last two types of errors mentioned above.

\section{Cross-lingual Experiments}
This section introduces the cross-lingual named entity recognition work in the paper. We first explain our motivation and illustrate the overview of the dataset translation process, followed by a description of the experiment setup. Table \ref{en-zh} and Table \ref{zh-en} present the results of the cross-lingual experiments, which highlight the effectiveness of cross-lingual approaches in improving NER performance.

\subsection{Motivation}

Despite the inherent disparities between Chinese and English, such as the absence of separators in Chinese compared to English, as well as differences in expressive conventions, we have identified shared syntactic structures between the two languages. These commonalities, such as the consistent use of names as subjects or objects in sentences, and the placement of locations as adverbial elements with prepositions, provide an opportunity to enhance the accuracy of NER. Similarly, organizations often function as subjects or adverbial elements in both languages. Based on this, we conjecture that Chinese and English can complement each other to enhance the performance of NER models. In addition, due to the deficiency of the Chinese multimodal NER dataset, there has been limited research on multimodal cross-lingual named entity recognition between Chinese and English. Consequently, we conduct cross-lingual experiments on our constructed CMNER dataset and Twitter2015 to validate our hypothesis.

\subsection{Dataset and Preprocessing}

For the English dataset, we employ Twitter2015 to conduct cross-lingual experiments with the Chinese CMNER. Motivated by \cite{fei2020cross}, we adopt a translation-based method to address this task, which requests high-quality training datasets for the target language by leveraging annotated datasets in the source language. Following the approach outlined in \cite{zhou2022conner}, we use an alignment-free way to translate the dataset into a different language while keeping entity spans fixed. In specific, we first replace the entity span with the placeholder ``SPAN'' in the raw text and then translate the sentence, ensuring that the placeholder ``SPAN'' remains unaltered during the translation process. Subsequently, we substitute the placeholder with the translation of the concrete entity span to obtain the complete expression of the raw text in the target language. Finally, we map the labels from the original dataset to the translated one, yielding a high-quality annotated dataset in the target language.

\subsection{Experiment Setup}

To verify our hypothesis, We design a dual set of experiments. One set involves English as the source language and Chinese as the target language, while the other set has Chinese as the source language and English as the target language. We denote the former as ``en→zh'' and the latter as ``zh→en''. In both cases, we conduct comprehensive experiments using both ACN and UMT. For each direction, we consider three training configurations: using only the source corpus(SRC), using only the translated target corpus(TGT), and using a combined corpus of both source and target(SRC\&TGT). Due to the distinct posting habits of users on Twitter and Weibo, the datasets exhibit unique characteristics in terms of content. For instance, Weibo posts tend to be longer, whereas tweets are comparatively shorter. Consequently, after training on the source language dataset, the model requires further fine-tuning on the target language dataset to adapt to these differences. To be specific, we employ the CMNER dataset for fine-tuning in the ``en→zh'' experiments and Twitter2015 in the ``zh→en'' experiments.

For multilingual representations in ACN, we use MUSE \cite{conneau2017word} to align all monolingual word embeddings into a universal space. For UMT, we employ the officially released multilingual BERT (base, cased version). The remaining parameters are kept consistent with those in Section \ref{sec:setup}. As recommended by \cite{wu2019beto}, when using multilingual BERT, we freeze the bottom 3 layers to maximize its performance for MNER.

\subsection{Results and Analysis}

% Please add the following required packages to your document preamble:
% \usepackage{multirow}
\begin{table}[!t]
\centering
\small
\begin{tabular}{ccccc}
\hline

\hline
\multicolumn{1}{c}{Model}  &Train        & P          & R    & F1             \\
% \cmidrule(r){3-3}\cmidrule(r){4-4}\cmidrule(r){5-5}
\hline
\multirow{4}{*}{ACN} & -        & 76.03          & 73.07          & 74.52          \\
                     & SRC      & 78.93          & 72.54          & 75.60          \\
                     & TGT      & 82.31          & 70.97          & 76.22          \\
                     & SRC\&TGT & 79.43          & 74.21          & 76.73          \\
                     \hline
\multirow{4}{*}{UMT} & -        & 87.08          & 90.16          & 88.60          \\
                     & SRC      & 86.97          & 90.47          & 88.69          \\
                     & TGT      & 87.88          & 90.03          & 88.95          \\
                     & SRC\&TGT & \textbf{88.05} & \textbf{90.53} & \textbf{89.27} \\
                     \hline

\hline
\end{tabular}
\caption{Results of the ``en→zh'' experiments.}
\label{en-zh}
\end{table}

% Please add the following required packages to your document preamble:
% \usepackage{multirow}
\begin{table}[]
\centering
\small
\begin{tabular}{ccccc}
\hline

\hline
\multicolumn{1}{c}{Model}  &Train      & P              & R              & F1             \\
\hline
\multirow{4}{*}{ACN}    & -         & 63.72          & 63.89          & 63.81          \\
                        & SRC       & 66.48          & 61.71          & 64.00          \\
                        & TGT       & 68.71          & 63.15          & 65.81          \\
                        & SRC\&TGT  & 69.98          & 62.65          & 66.11          \\
                        \hline
\multicolumn{2}{c}{ATTR-MMKG-MNER}  & 74.78          & 71.82          & 73.27          \\
\multicolumn{2}{c}{UMGF}            & 74.49          & 75.21          & 74.85          \\
\multicolumn{2}{c}{MAF}             & 71.86          & 75.10          & 73.42          \\
\multicolumn{2}{c}{MNER-QG}         & 77.43          & 72.15          & 74.70          \\
\multicolumn{2}{c}{MNER-QG(Oracle)} & \textbf{77.76} & 72.31          & 74.94          \\
\hline
\multirow{4}{*}{UMT}    & -         & 71.67          & 75.23          & 73.41          \\
                        & SRC       & 73.26          & 76.44          & 74.81          \\
                        & TGT       & 73.23          & 76.65          & 74.90          \\
                        & SRC\&TGT  & 73.82          & \textbf{76.93} & \textbf{75.34} \\
                        \hline

                        \hline
\end{tabular}
\caption{Results of the ``zh→en'' experiments.}
\label{zh-en}
\end{table}

We utilize precision, recall, and F1 scores for all entity types as the evaluation metrics. The results of the "en→zh" experiments are presented in Table \ref{en-zh}, while those of the "zh→en" experiments are shown in Table \ref{zh-en}. In the tables, ``-'' denotes the model is directly trained on the target dataset while ``SRC'' indicates the model is initially trained on the source language corpus and then fine-tuned on the target dataset. ``TGT'' and ``SRC\&TGT'' signify that the model is first trained on the translated target corpus and the mixture corpus, respectively.

In Table \ref{en-zh}, the top four rows present the results for ACN, while the bottom four rows display the results for UMT. Notably, We observe a consistent improvement in the performance of both models, which affirms the efficacy of incorporating English data in Chinese NER tasks. When utilizing the mixed corpus for training, both two models achieve their highest F1 scores. ACN exhibits a greater improvement, increasing from 74.52 to 76.73, a notable gain of 2.21 points. Whereas, UMT shows a more modest advancement, with an increase of 0.67 points, from 88.60 to 89.27. This discrepancy can be attributed to the inherent capabilities of the models. UMT already possesses a higher baseline performance, thus making further gains more challenging to achieve.

Table \ref{zh-en} shows the results of the ``zh→en'' experiments conducted on Twitter2015. In addition to our proposed approaches, the table also displays the results of existing methods used as baselines. The first four rows detail the results for ACN, followed by five rows presenting various baselines from \cite{jia2023mner}, and the last four lines record the results for UMT. For context, ATTR-MMKG-MNER \cite{chen2021multimodal} incorporates image attributes and image knowledge into MNER, while UMGF \cite{zhang2021multi}  stack multiple graph-based multi-modal fusion layers to learn representations. MAF \cite{xu2022maf} proposes a general matching and alignment framework for MNER, and MNER-QG \cite{jia2023mner} utilizes a machine reading comprehension framework with query grounding to address the task. The results illustrate that when trained on the mixed corpus and subsequently fine-tuned, UMT achieves the highest F1 scores among all models, reaching 75.34. Similar to the trends observed in the ``en→zh'' experiments, both ACN and UMT demonstrate a gradual improvement in performance. ACN shows a larger improvement, increasing by 2.3 points from 63.81 to 66.11, while UMT gains an improvement of 1.93 points, resulting in an F1 score of 75.34. These statistics further verify that utilizing Chinese as the source language can enhance the performance of English NER models as well.

The comprehensive results of the cross-lingual experiments in both two directions illustrate that Chinese and English can reciprocally augment the performance of NER models and thereby validate our initial hypothesis.

\section{Conclusion and Future Work}
In this paper, we propose CMNER, a manually annotated Chinese multimodal NER dataset sourced from social media. CMNER addresses the scarcity of Chinese multimodal corpora and encompasses diverse topics and entity categories. We implement baselines and conduct a series of cross-lingual experiments, the results of which indicate that CMNER is a challenging corpus with substantial underlying research prospects. In the future, we intend to (1) construct a more sophisticated multimodal model to fully leverage implicit information in multiple images (2) take advantage of images as a bridge for cross-lingual MNER tasks, and (3) extend the model's application to a broader range of scenarios.

\section{Bibliographical References}

\bibliographystyle{lrec-coling2024-natbib}
\bibliography{lrec-coling2024-example}

% \section{Language Resource References}
% \label{lr:ref}
\bibliographystylelanguageresource{lrec-coling2024-natbib}
\bibliographylanguageresource{languageresource}

\end{document}